\pdfoutput=1
\documentclass[11pt]{article}

\usepackage{ACL2023}

\usepackage{times}
\usepackage{latexsym}
\usepackage{amssymb}

\usepackage[T1]{fontenc}
\usepackage[utf8]{inputenc}

\usepackage{microtype}

\usepackage{inconsolata}
\usepackage{xspace}
\usepackage{tcolorbox}
\usepackage{soul}
\usepackage{setspace}

\usepackage{booktabs}
\usepackage{graphicx}
\usepackage{xspace}
\usepackage{makecell}
\usepackage[normalem]{ulem}
\usepackage{tabularx}

\newcommand\clearrow{\global\let\rowmac\relax}
\clearrow

\definecolor{forestgreen}{RGB}{34, 139, 34}
\definecolor{olive}{RGB}{128, 128, 0}
\definecolor{deepskyblue}{RGB}{0, 191, 255}
\definecolor{bbrown}{RGB}{165, 42, 42}

\usepackage{listings}
\usepackage{color}

\definecolor{dkgreen}{rgb}{0,0.6,0}
\definecolor{gray}{rgb}{0.5,0.5,0.5}
\definecolor{mauve}{rgb}{0.58,0,0.82}

\definecolor{codegreen}{rgb}{0,0.6,0}
\definecolor{codegray}{rgb}{0.5,0.5,0.5}
\definecolor{codepurple}{rgb}{0.58,0,0.82}
\definecolor{backcolour}{rgb}{0.95,0.95,0.92}

\lstdefinestyle{mystyle}{
    backgroundcolor=\color{backcolour},   
    commentstyle=\color{codegreen},
    keywordstyle=\color{magenta},
    numberstyle=\tiny\color{codegray},
    stringstyle=\color{codepurple},
    basicstyle=\ttfamily\footnotesize,
    breakatwhitespace=false,         
    breaklines=true,                 
    captionpos=b,                    
    keepspaces=true,                 
    numbers=left,                    
    numbersep=5pt,                  
    showspaces=false,                
    showstringspaces=false,
    showtabs=false,                  
    tabsize=2
}

\newcommand{\sectionRef}[1]{\S\ref{#1}}
\newcommand{\figureRef}[1]{Figure~\ref{#1}}

\newcommand{\unitxt}[0]{\texttt{Unitxt}\xspace}
\newcommand{\catalog}[0]{\texttt{Catalog}\xspace}

\usepackage{xcolor}

\definecolor{template_color}{HTML}{EC9BFA}  
\definecolor{format_color}{HTML}{D9E0F5} 
\definecolor{task_color}{HTML}{7A81FF} 
\definecolor{resource_color}{HTML}{EFD614} 
\definecolor{prompt_color}{HTML}{FE6100} 
\definecolor{extension_color}{HTML}{000000} 

\newcommand{\coloredsquare}[1]{\textcolor{#1}{$\blacksquare$}}

\title{\texttt{Unitxt}: Flexible, Shareable and Reusable \\
Data Preparation and Evaluation for Generative AI }
\author{\normalsize
Elron Bandel \qquad Yotam Perlitz \qquad Elad Venezian \qquad Roni Friedman-Melamed \\
\textbf{ \normalsize Ofir Arviv} \qquad \textbf{ \normalsize Matan Orbach} \qquad
\textbf{ \normalsize Shachar Don-Yehyia} \qquad \textbf{ \normalsize Dafna Sheinwald}\\ 
\textbf{ \normalsize Ariel Gera} \qquad  \textbf{ \normalsize Leshem Choshen} \qquad 
\textbf{ \normalsize Michal Shmueli-Scheuer} \qquad \textbf{ \normalsize Yoav Katz} \\
IBM Research \\
\texttt{elron.bandel@ibm.com}}

\begin{document}
\maketitle
\begin{abstract}

In the dynamic landscape of generative NLP, traditional text processing pipelines limit research flexibility and reproducibility,  as they are tailored to specific dataset, task, and model combinations. 
The escalating complexity, involving system prompts, model-specific formats, instructions, and more, calls for a shift to a structured, modular, and customizable solution.
Addressing this need, we present \unitxt, an innovative library for customizable textual data preparation and evaluation tailored to generative language models. \unitxt natively integrates with common libraries like HuggingFace and LM-eval-harness and deconstructs processing flows into modular components, enabling easy customization and sharing between practitioners. 
These components encompass model-specific formats, task prompts, and many other comprehensive dataset processing definitions. The \unitxt \texttt{Catalog} centralizes these components, fostering collaboration and exploration in modern textual data workflows. Beyond being a tool, \unitxt is a community-driven platform, empowering users to build, share, and advance their pipelines collaboratively. Join the \unitxt community;\\
Project: \url{https://github.com/IBM/unitxt}.\\
UI: \url{https://bit.ly/unitxt-explore} \\
Video: \url{https://bit.ly/unitxt-video} 

\end{abstract}

\section{Introduction}

Textual data processing has always been at the heart of NLP, but in the current landscape it has taken on new roles. 
A prominent one comes from LLMs' role as general interfaces, that receive an example, but also the task they should perform, general system instruction and other specifications, all in natural language. 
Thus, the inputs -- or \textit{prompts} -- that a model receives now consist of many components, that can be combined in different ways: task instructions \citep{wei2022finetuned}, in-context demonstrations \citep{ICL}, system prompts and more. 
At the same time, for text generation models, model outputs are themselves rich textual data, and thus can be processed and evaluated with a range of different approaches and paradigms. 
Therefore, textual data processing for LLMs is growing increasingly complex. It incorporates a large number of non-trivial design choices and parameters, which pose new challenges for maintaining flexibility and reproducibility in LLM research.

Broadly, research in computer science, and in particular within NLP, thrives on that combination of flexibility and reproducibility. 
On the one hand, it should be simple to try new ideas: to compare different approaches, choose parameters, and easily switch out one workflow or architecture with another. 
On the other hand, the results of these explorations must be shared in such a way that others are able to -- and crucially, are likely to -- reproduce and try them. 
To enable the above, code reuse, a well-defined API and ease of use are pivotal, ensuring reproducibility and applicability in practice. How such traits allow for widespread adoption is epitomized by the Hugging Face transformers library~\citep{wolf-etal-2020-transformers}. 
Today, a modest set of hyperparameters is sufficient to reproduce a training or inference workflow. This has had an undeniable and dramatic impact on the ability to make progress in the field.

Such is not the case, however, for textual data pipelines. Unfortunately, data-preparation for LLMs has no standards, Processing model inputs or outputs of the same data often comes with rewriting the code, leading to mismatches in reported values \citep{post2018sacreblue}, unanswerable examples and hidden bugs \citep{DropBlog} and general time waste. 
Crucially, the additional components beyond traditional processing, such as in-context demonstrations, have no canonical API. 
% why is this bad
This prevents fair comparisons between different studies, discourages exploring combinations, hinders integrating a particular approach (say, a new type of system prompt) into an existing NLP system, and prevents major scale-ups in terms of datasets, tasks and metrics. 

\begin{figure*}
    \centering
    \includegraphics[width=1\linewidth]{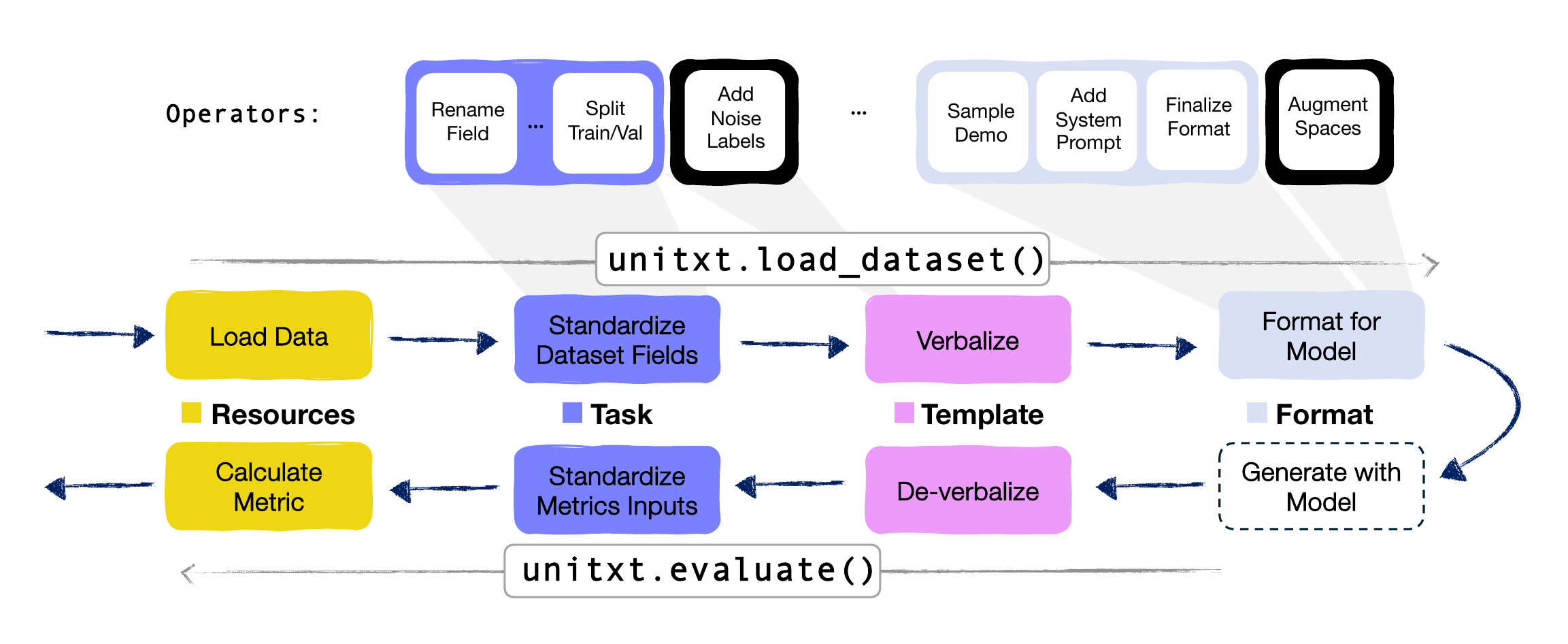}
    \caption{\unitxt flow: The upper section illustrates the data-preparation pipeline \S\ref{data_preperation_pipeline}, encompassing raw dataset loading, standardization according to the task interface, verbalization using templates, and application of formatting. 
    The lower section showcases the evaluation pipeline \S\ref{evaluation_pipline}, involving de-verbalization operations and output standardization before performance evaluation with task-defined metrics.
    All components are described in \sectionRef{format}.}
    \label{fig:unitxt-flow}
\end{figure*}

% what do we show here
To address these gaps, we introduce a new collaborative framework for unified textual data processing named \unitxt.  
%MO To address these gaps, we introduce a new Python library named \textit{unitxt} which serves as a collaborative framework for unified textual data processing. 
% elaborate on what unitext does
% many (most maybe?) things are already supported
This new Python library supports multilingual textual data processing through flexible pipelines called \textit{recipes}. 
A recipe (see \S\ref{recipe} and examples in \S\ref{sec:tour}) is a sequence of textual data processing operators, including, among others, 
operators that load data, pre-process it, handle the preparation of different parts of a prompt, or evaluate model predictions (see \figureRef{fig:unitxt-flow}).
%MO operators handling the preparation of different parts of a prompt: verbalizing inputs from raw data, adding system prompts, integrating in-context demonstrations, or adjusting input to the format of a specific model (see \figureRef{fig:verbalization}).
%MO Operators also take care of the data preparation and evaluation of model predictions through a diverse set of mertics.

Aiming for reuse, \unitxt ships with a catalog containing a wide variety of pre-defined recipes for various tasks. 
These are all based on a diverse set of built-in operators that are also shared in the catalog.
Having a centralized location for these components, where anyone can add new ingredients (such as recipes or operators), or share existing ones, fosters collaboration, transparency and reproducibility.

%MO To enable collaboration, transparency and reproducibility, unitxt is build upon a catalog of crowd-sourced recipes and textual-processing operators. Anyone can add new recipes or share existing ones. 

%MO This new Python library supports a wide variety of \textbf{X} pre-existing \dafna{pre-defined?} textual data processing pipelines (denoted here as \textit{recipes}), this number represents a great opportunity for scale at the minimal cost of integration.
% changing the recepies is easy due to the modular charecters

As fitting a Recipe, the modularity of \unitxt enables mixing and matching of ingredients to create new recipes. 
This ability to mix and match ingredients enables \unitxt to support $100$K+ recipe configurations, allowing users to experiment with a large set of such recipes by to obtain multiple configurations of tasks, datasets and new formatting (see \S\ref{sec:tour} for example). 

Changing libraries is always a nuisance; therefore, \unitxt is designed to seamlessly integrate with preexisting code, offering a hassle-free experience without even needing a pip install. 
For instance, \unitxt can load HuggingFace datasets and produce outputs that adhere to the same format, allowing it to integrate seamlessly with other parts of your codebase (\S\ref{data_preperation_pipeline}). Demonstrating this, incorporating \unitxt, with all its tasks, datasets, templates and metrics into LM-eval-harness~\cite{eval-harness} required only 30 lines of code, while preserving the current API and  ensuring a smooth transition and compatibility with existing workflows (App.~\ref{ap:lmeval}).

\unitxt, an open-source library, is under active development by IBM and the community. 
The code and documentation are available on GitHub at: \url{https://github.com/IBM/unitxt}, the UI, at \url{https://bit.ly/unitxt-explore} while the demo video is at \url{https://bit.ly/unitxt-video}.

%MO: not sure this should be so early. at this point the reader doesn't have an idea, i think, of what unitxt is. so perhaps that should be before the use cases.
\section{Use cases}

\textbf{\unitxt for evaluation}:  
The increasing capabilities of LLMs require
%MO prompted the need for 
evaluation frameworks that test models over an unprecedented number of datasets, tasks and configurations~\cite{liang2022holistic, eval-harness, 2023opencompass}.
\unitxt can serve as the backbone of such evaluation efforts, by supporting easy changes across multiple important axes, including tasks, languages, prompt structure (e.g. instructions, verbalizations, etc.), augmentation robustness and more. 
Moreover, 
with the \unitxt \catalog, %the collaborative nature of \unitxt allows for, 
different distinct projects can share their full evaluation pipelines, making their data-preparation and evaluation metrics reproducible.
%MO through the catalog holding hundreds of internal dataset-task combinations.
% \matan{maybe add a concrete list of tasks that being used unitxt within IBM? and add the number of runs in a typical unitxt-powered evaluation in IBM? its hundreds or thousands of runs, no? per new model?}

\textbf{\unitxt for training}: 
Modern LLM training frameworks have extensive data requirements to attain state-of-the-art performance. Multiple datasets across diverse domains and languages need to be leveraged to impart broad capabilities;
Various prompt formulations and \emph{verbalizations} are necessary to enable instruction-following, where verbalizations are the final text form. 
However, combining heterogeneous data sources and textual representations poses significant engineering challenges. Without a common underlying framework, data augmentation, multitask learning and few-shot tuning become prohibitively complex. 
This is where \unitxt steps in, as an indispensable data backend.

\unitxt enables seamless fusion of diverse datasets. 
Moreover, the standardized format also facilitates changes to the datasets, dynamic prompt generation, data augmentations and model-specific format, to name just a few. 
By handling the data wrangling complexity, \unitxt empowers researchers to focus on creating performant, robust and safe LLMs. 

For both evaluation and training, \unitxt \emph{has already been adopted} as a core utility for LLMs in IBM by multiple teams working on various NLP tasks, including classification, extraction, summarization, generation, question answering, code, biases and more. In total,  the open source catalog contains more than $100$K possible pipeline configurations.

% \michal{there is some imbalance between the two use-cases, where it's clear that the first one is extensively using unitxt, whereas the second one is more theoretical. maybe we need to take the specific IBM from the evaluation, and from the beginning. First present two use cases, and then summarize by saying that IBM... with some statistics as matan suggested. So it's not specific only for evaluation}

\section{Unitxt: Library Tour}\label{sec:tour}

To introduce unitxt, we begin with a tour of the library, and specifically, with the creation of a recipe. 
A recipe contains all the data-processing and metric configurations needed, including the data, task, template and formatting (see details in \S\ref{sec:design}). Here we define a recipe that loads the STS-B dataset for a sentence similarity task: 

\begin{lstlisting}
recipe = """
    card=cards.stsb, # dataset info card
    template=templates.text_similarity,
    sys_prompt=prompts.helpful,
    format=formats.user_agent,
    num_demos=1
"""
\end{lstlisting}
With a recipe, a concrete dataset can be loaded:
\begin{lstlisting}[language=Python,numbers=none]
dataset = unitxt.load_dataset(recipe)
\end{lstlisting}
Importantly, every data instance in a dataset loaded with a unitxt recipe contains a fully prepared source text, which can be directly passed as input to the model. 
For example, here is such source text for one sentence-similarity data instance, integrated with three formatting decisions, a ``helpful model'' system-prompt, a user-agent response schema and one demonstration:

\begin{lstlisting}[backgroundcolor=\color{white},numbers=none]
[System] you are helpful model [/System]
[User]: for the following texts rank the 
        similarity between 1 to 5.
        Text 1: "i love ice cream"
        Text 2: "i like ice cream"
[Agent]: 4.8
[User]: Text 1: "i hate pizza"
        Text 2: "i like pizza"
[Agent]:
\end{lstlisting}

Loading a dataset with a unitxt recipe also adds a  metric-ready target text (created from the original target) to each data instance.
To Evaluate the model's textual predictions, we call:
\begin{lstlisting}[language=Python,numbers=none]
results = unitxt.evaluate(
   dataset,
   predictions=predictions, 
)
\end{lstlisting}
The evaluation results are a dictionary of task defined metric names and the values computed for them.

\section{Design}\label{sec:design}

% what we aim in this section
In this section we outline the design of \unitxt.  
\unitxt processes data by applying a modular sequence of operators, which are segmented into $5$ key ingredients (\S\ref{ingredients}) color-coded as in Fig.~\ref{fig:unitxt-flow}: 
\coloredsquare{resource_color}~Resources, 
\coloredsquare{task_color}~Task, 
\coloredsquare{template_color}~Template, 
\coloredsquare{format_color}~Format 
and \coloredsquare{extension_color}~Extensions. 
These ingredients are then used to build the \textbf{data preparation} (\S\ref{data_preperation_pipeline}) and \textbf{evaluation} (\S\ref{evaluation_pipline}) pipelines. 

\subsection{\unitxt Building Blocks}
\label{subsec:building_blocks}
When loading a dataset (as demonstrated in \S\ref{sec:tour}), the \unitxt ingredients are retrieved based on a \textit{Data-Task Card} and a \textit{Recipe}.

\paragraph{\coloredsquare{resource_color}\coloredsquare{task_color} Data-Task Card}
\label{subsec:taskaset}
Defines how raw data (inputs and targets) are standardized for a certain task.
Typically, this includes data wrangling actions, e.g. renaming fields, filtering data instances, modifying values, train/test/val splitting etc. 
It also describes the resource from which the data is loaded.

\paragraph{\coloredsquare{resource_color}\coloredsquare{task_color}\coloredsquare{template_color}\coloredsquare{format_color}\coloredsquare{extension_color} Recipe} \label{recipe}
A \emph{Recipe} holds a complete specification of a \unitxt pipeline: including the Resources, Task, Template, Format and Extensions.

\subsection{\unitxt Ingredients} \label{ingredients}

\paragraph{\coloredsquare{resource_color} Resources}
Raw data and metrics are external resources utilized by \unitxt.
\unitxt implements several APIs for raw-data and metric loading (e.g., from Huggingface Hub, local files, and cloud storage).

\paragraph{\coloredsquare{task_color} Task}
A \unitxt \emph{Task} follows the formal definition of an NLP task, such as multi-label classification, named entity extraction, abstractive summarization or translation. 
A task is defined by its standard interface -- namely, input and output fields -- and by its evaluation metrics. 
Given a dataset, its contents are standardized into the fields defined by an appropriate task by a Data-Task Card (\S\ref{subsec:building_blocks}).

As an example of a defined task, consider sentence similarity: 
it has two input fields (named \texttt{``sentence1''} and, \texttt{``sentence2''}), one output field (named \texttt{``label''}) and the conventional metric is Spearman correlation~\cite{spearman}. 

\paragraph{\coloredsquare{template_color} Template}
\label{template}

\begin{figure}[t]
    \includegraphics[width=\columnwidth]{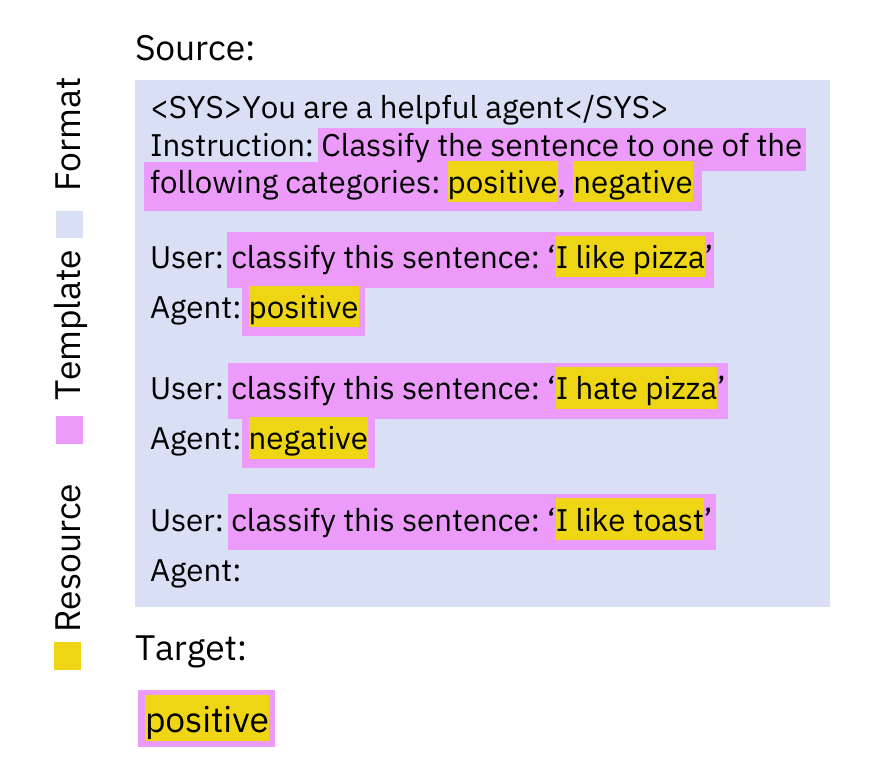}
% \end{figure*}
    \caption{Illustration of the data preparation pipeline (\sectionRef{data_preperation_pipeline}), depicting the transformation from raw data and formatting specifications to the final text output. Components include Resources (raw data), Format (model-specific formatting requirements), and Template (verbalization).}
    \label{fig:verbalization}
\end{figure}

A \unitxt \emph{Template} defines the verbalizations to be applied to the inputs and targets, 
as well as the de-verbalization operations over the model predictions. 
For example, in Fig~\ref{fig:verbalization}, applying the template to \texttt{I like toast} verbalizes it into \texttt{classify the sentence: ``I like toast''}.

In the other direction, template de-verbalization involves two steps. First, a general standardization of the output texts: taking only the first non-empty line of a model's predictions, lowercasing, stripping whitespaces, etc. The second step standardizes the output to the specific task at-hand.
%MO which is  imperative for a fair and consistent evaluation of different templates defining varying styles of response.
For example, in Sentence Similarity, a prediction may be a quantized float number outputted as a string (e.g ``2.43''), or a verbally expressed numeric expression (e.g ``two and a half''). This depends on the verbalization defined by the template and the in-context demonstrations it constructs.  
%MO To conform with the evaluation metrics that follow, all kinds 
Both types of outputs should be standardized   before evaluation begins -- e.g. to a float for sentence similarity. 
Having the de-verbalization steps defined within the template enables templates reuse across different models and datasets.

Crucially, in contrast to existing solutions (e.g., \citealp{bach2022promptsource}) the templates, datasets and tasks in \unitxt are not exclusively tied. Each task can harness multiple templates and a template can be used for different datasets. 
Thus, the modularity of \unitxt allows mixing and matching, significantly enhancing re-usability and flexibility.

\paragraph{\coloredsquare{format_color} Format}
\label{format}

A \unitxt \emph{Format} defines a set of extra formatting requirements, unrelated to the underlying data or task, including those pertaining to system prompts, special tokens or user/agent prefixes, and in-context demonstrations. 
Continuing the example from \figureRef{fig:verbalization}, the \unitxt format receives the text produced by the template \texttt{classify the sentence: ``I like toast''}, and adds the system prompt \texttt{<SYS>You are a helpful agent</SYS>}, the Instruction-User-Agent schema cues, and the two presented demonstrations.

\paragraph{\coloredsquare{extension_color} Extensions} \label{extensions}
\unitxt supports \textit{Extensions} such as \textit{input-augmentation} (for example, adding random whitespace, introducing spelling mistakes, or replacing words with their synonyms) or \textit{label-noising} (replaces the labels in the demonstrations randomly from a list of options). 
Such extensions can be added anywhere in the data-preparation pipeline between any two operators, depending on the desired logic (see Fig.~\ref{fig:unitxt-flow}). \unitxt supports the addition of custom extensions to the \catalog. Each extension is an independent unit, reusable across different datasets and tasks, templates and formats. 

\subsection{\unitxt \catalog}

All \unitxt artifacts -- recipes, data-task cards, templates, pre-processing operators, formats and metrics -- are stored in the \unitxt \catalog. 
% there are many such catalogs
In addition to the open-source catalog, that can be found in the documentation, users can choose to define a private catalog. 
This enables teams and organizations to harness the open \unitxt \catalog while upholding organizational requirements for additional proprietary artifacts.

\subsection{\unitxt Pipelines}
\subsubsection{Data Preparation Pipeline}
\label{data_preperation_pipeline}

% As seen in Fig~\ref{fig:unitxt-flow}, \unitxt data preparation includes loading the raw data from an external source, standardizing it to the task interface, using a template to verbalize each sample (adding instruction etc.), and then formatting the verbalized samples (e.g., by adding system prompts, special tokens and in-context learning examples).

% \subsubsection{\coloredsquare{resource_color}\coloredsquare{task_color} Loading and Standardizing to Task}

% loading
The data preparation pipeline (top part ot Fig.~\ref{fig:unitxt-flow}) begins with
% MO with the raw dataset loading. 
% task
standardizing the raw data 
%MO The data is then standardized 
into the task interface, as defined in the data-task card (\S\ref{subsec:taskaset}).
% template
%MO Once standardized according to the task interface, 
The examples are then verbalized by the template, and the format 
%MO Closing the data-preparation stage is the application of formatting over the verbalized samples. Such formatting may include adding
operator applies system prompts, special tokens and in-context learning examples (\S\ref{template}), as illustrated in \figureRef{fig:verbalization}.
% it is a HF datset
To maintain compatibility, the output of this pipeline is an HF dataset, that can be saved or pushed to the hub.

\begin{figure*}
    \centering
    \includegraphics[width=0.95\textwidth]{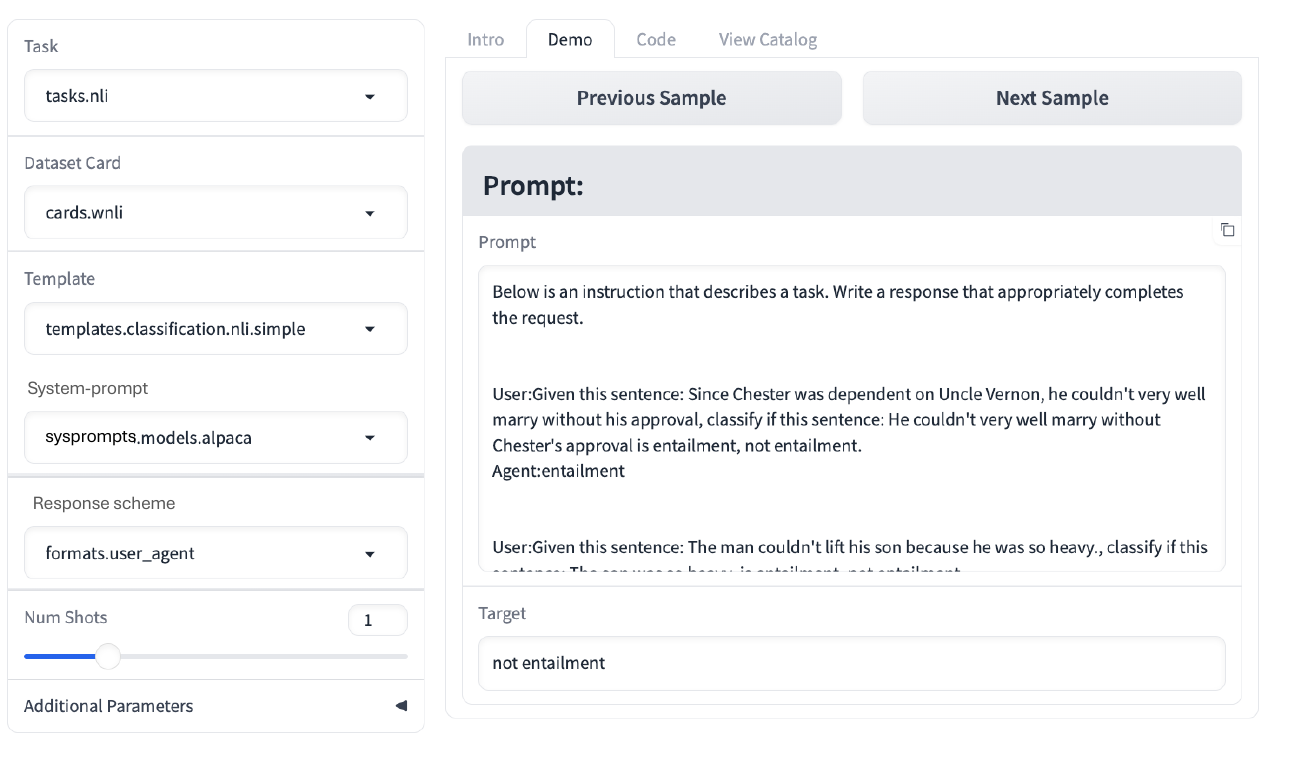}
    \caption{Exploration UI showcasing configuration options for model input creation on the left, including parameters such as task, dataset card, template, system-prompt, response-schema, number of examples, and optional augmentations. The resulting model input is displayed in the prompt window.} 
    \label{fig:ui}
\end{figure*}

\subsubsection{Evaluation Pipeline}
\label{evaluation_pipline}

The evaluation pipeline (bottom part of Fig.~\ref{fig:unitxt-flow})
is responsible for producing a list of evaluation scores that reflect model performance. It includes a de-verbalization of the model outputs (as defined in the template, see \sectionRef{template}), and a computation of performance 
by the metrics defined in the task. 
The standardization of the task interface, namely, having fixed names and types for its input and output fields, allows the use of any metric that accept such fields as input.  
In addition to the computed evaluation scores, \unitxt metrics supports a built in mechanism for confidence interval reporting, using statistical bootstrap ~\cite{perlitz2023efficient}.

\section{\unitxt UI:  Explore \& Preview  }

The objective of the user interface is to guide users through the essential steps of recipe creation, illustrated with pertinent examples. Additionally, it allows for catalog exploration.
The UI complements the experience with the option to execute the examples on some pre-set model (e.g., flan-t5-base), get the predictions and associated scores.

The interaction entry point is the tasks. 
Upon clicking, the tasks taxonomy is presented, and the users have the option to choose the applicable task type. Selecting a task results in showing only the relevant datasets and templates. 
Once the user selects a dataset, and a template, and presses ``Generate Prompts'' a random example enhanced with the template is loaded.
% Next, the user selects one of the templates, and at this point, an initial input appears in the demonstration view. 
If the user wants to augment the input with system prompt, or response-schema those will be instantly added when opted for. 
As in-context learning evaluations are supported, the user can select the preferred number of shots.  
Once satisfied with the example, the user has the option to proceed with executing it on a model, wherein the predictions and corresponding scores will be displayed for this specific example. Further, going to the code tab, the user can copy the associated code into a notebook and run. Users have the option to explore various examples, enhancing their comprehension and confidence in the chosen configuration.

\section{Related work}

Standardized data processing for evaluation and training has been a longstanding need in the NLP community and has been repeatedly addressed in the past.
% What are they
Datasets \cite{lhoest-etal-2021-datasets} and Evaluate\footnote{\url{https://github.com/huggingface/evaluate/}} 
 are community-driven libraries, providing a standardized interface to diverse corpora and metrics, as well as supporting many data processing operations. 
% What dont they do
These packages, however, fall short of providing a standardized, shareable and reproducible framework to cast the raw data into textual prompts and cast them back from text to a metric digestible format. 
% This is bad for reproducability
The lack of such a framework hinders reproducibility, as often slight variations in ad-hoc text processing code may yield significantly different evaluation scores. 
% This is bad for scaling up
Moreover, it also prevents users from easily scaling up their experiment, as each task and dataset often requires specific code for processing and evaluation.
% the absence of componentization of these parts of the text-to-text training/evaluation pipeline prevents users from easily scaling up their use for multiple, diverse tasks and datasets.
\unitxt builds on top of these frameworks, harnessing them as resources (\S\ref{subsec:building_blocks}) to produce a full data-preparation and evaluation framework.

While several existing frameworks have contributed to data pipeline management workflows, a common drawback, for those we are aware of, is the absence of a well-defined and flexible modularity in their design, such as the ability to define specific components for system prompts, task instructions and model-specific formats. This absence of clearly defined components makes it challenging to share and customize such pipelines effectively, across different datasets and tasks.

Like \unitxt, Tasksource~\cite{sileo2023tasksource} supplies tools for consistent preprocessing over different datasets, simplifying their usage. However, it is primarily designed for discriminative tasks, uses fixed formats and lacks a modular design that enables sharing, mixing and matching, and overall flexibility in processing steps.
Promptsource~\cite{bach2022promptsource} focuses on making and sharing natural language prompts but doesn't handle other types of data processing. Each prompt is tied to just one dataset, making it hard to reuse and share. Furthermore, prompts aren't split into system, instruction, and format parts, limiting options for flexibility and reuse.
SeqIO~\cite{roberts2022t5x}, offers task-based pipelines encompassing pre-processing, post-processing, and handling metrics. However, a structured breakdown of these processing steps is absent, limiting the creation of shareable catalogs within the community. In this framework, each process is a generic function and specialized steps are missing, like those designed for system prompts.

A different branch of solutions are language model evaluation frameworks such as OpenCompass~\cite{2023opencompass}, HELM~\cite{liang2022holistic} and LM-eval-harness~\cite{eval-harness} also implement their own standardized data processing pipelines in order to obtain verbalized prompts for LMs. These, however, are highly coupled with the inference engine and cannot be used as standalone data-processing pipelines or integrated into other code bases.

\section{Conclusion}

In this paper, we have introduced \unitxt, an open-source Python library aimed at unifying textual data processing pipelines for large language models. \unitxt provides a modular, flexible framework that enables mixing and matching of various pipeline components like loaders, templates, formats and metrics. \unitxt key capabilities are, standardization, flexibility, collaboration and scale.

\unitxt has already been successfully deployed for large language model evaluation and training within IBM. As the library matures through open-source community involvement, we hope its adoption will grow to push the frontiers of textual data processing for LLMs. 
We believe \unitxt has the potential to significantly impact research and development of large language models by unifying textual data processing. Through its emphasis on flexibility, reproducibility and collaboration, unitxt can help drive progress towards more capable, safer and trustworthy LLMs.

\section{Limitations}

While unitxt makes significant progress towards unified textual data processing for LLMs, some limitations still remain:

\begin{itemize}

\item The \unitxt \catalog, while already substantial in coverage, needs expansion to encompass more datasets, languages, and niche tasks. Community contributions will be key to enhancing catalog diversity.

\item Coverage of evalution metrics, especially for generative tasks, needs improvement. We plan to incorporate more reference-free and LLM-based metrics going forward.

\item Training data augmentation abilities, while flexible currently, can be expanded further with techniques like back-translation for multilinguality.

\item While using \unitxt recipes is as simple as specifying the recipe ingrediants, adding new datasets or operators requires learning the \unitxt operator language. Additional documentation, examples and IDE support could help alleviate this.

\end{itemize}

Addressing these limitations through open-source community involvement is the major focus going forward. By tapping into collective expertise, we envision unitxt becoming an indispensable textual data processing backbone for the responsible development, evaluation and deployment of large language models.

\bibliography{custom}
\bibliographystyle{acl_natbib}

\appendix
\section{LM Eval Harness Integration}
\label{ap:lmeval}

% \section{Case study} \label{app:usecase-lmeval}

LM-eval-harness~\cite{eval-harness} is one of the most commonly used open source evaluation frameworks.  
It leverages a yaml-based declarative language which defines loading of the dataset, the dataset splits, the prompt used and the metrics in a single file for each task.  
Many tasks are supported, including multi-class classification, multiple choice question answering, and generation tasks.   
Unitxt was integrated into LM-eval-harness to extend LM-eval-harness to support new tasks and metrics that currently are not supported, including multi-label classification, named entity extraction, and target sentiment analysis.
%MO (should we mention, RAG and code)

Since \unitxt recipes can be loaded as standard HF datasets,  no code changes were required to add the \unitxt data preparation pipeline to LM-eval-harness. Adding a \unitxt recipe requires only one line change in a LM-eval-harness yaml (see \figureRef{fig:unitxt-lm-eval-harness} in Appendix).  Adding the \unitxt metrics required about 30 lines of code, to register the \unitxt metrics to the LM-eval-harness metrics registry.

\begin{figure*}
    \centering

\begin{lstlisting}
group: glue
task: unitxt_unfair_tos
dataset_path: unitxt/data
dataset_name: card=cards.unfair_tos,template_card_index=templates.classification.multi_label.default,format=formats.user_agent
output_type:  generate_until
training_split: train
validation_split: validation
doc_to_text: "{{source}}"
doc_to_target: target
generation_kwargs:
  until:
    - "</s>"
metric_list:
  - metric: unitxt_f1_micro_multi_label
metadata:
  version: 1.0
\end{lstlisting}
\caption{\textbf{Unitxt and LM-eval-harness integration}.  A Unitxt recipe can be integrated as an LM-eval-harness task, by setting the \emph{dataset\_path} (line 3) to \emph{unitxt/data} and the setting the recipe in the \emph{dataset\_name} (line 4).  Unitxt metrics can be used like any other metric (line 14).  }
\label{fig:unitxt-lm-eval-harness}
\end{figure*}

\end{document}